# Virtual Staining of Label-Free Tissue in Imaging Mass Spectrometry

Yijie Zhang[†,1,2,3], Luzhe Huang[†,1,2,3], Nir Pillar[1,2,3], Yuzhu Li[1,2,3], Lukasz G. Migas[4,5], Raf Van de Plas[4,5,8], Jeffrey M. Spraggins[4,6,7,8] and Aydogan Ozcan[*,1,2,3,9]

[1]Electrical and Computer Engineering Department, University of California, Los Angeles, CA, 90095, USA.

[2]Bioengineering Department, University of California, Los Angeles, CA, 90095, USA.

[3]California NanoSystems Institute (CNSI), University of California, Los Angeles, CA, 90095, USA.

[4]Mass Spectrometry Research Center, Vanderbilt University, Nashville, TN 37232, USA.

[5]Delft Center for Systems and Control, Delft University of Technology, 2628 Delft, The Netherlands

[6]Department of Cell and Developmental Biology, Vanderbilt University, Nashville, TN 37232, USA.

[7]Department of Chemistry, Vanderbilt University, Nashville, TN 37232, USA.

[8]Department of Biochemistry, Vanderbilt University, Nashville, TN 37232, USA.

[9]Department of Pathology, Microbiology, and Immunology, Vanderbilt University Medical Center, Nashville, TN 37232, USA.

[9]Department of Surgery, University of California, Los Angeles, CA, 90095, USA.

**\*Correspondence:** Aydogan Ozcan, ozcan@ucla.edu

[†] **Equal contributing authors**

# Abstract

Imaging mass spectrometry (IMS) is a powerful tool for untargeted, highly multiplexed molecular mapping of tissue in biomedical research. IMS offers a means of mapping the spatial distributions of molecular species in biological tissue with unparalleled chemical specificity and sensitivity. However, most IMS platforms are not able to achieve microscopy-level spatial resolution and lack cellular morphological contrast, necessitating subsequent histochemical staining, microscopic imaging and advanced image registration steps to enable molecular distributions to be linked to specific tissue features and cell types. Here, we present a virtual histological staining approach that enhances spatial resolution and digitally introduces cellular morphological contrast into mass spectrometry images of label-free human tissue using a diffusion model. Blind testing on human kidney tissue demonstrated that the virtually stained images of label-free samples closely match their histochemically stained counterparts (with Periodic Acid-Schiff staining), showing high concordance in identifying key renal pathology structures despite utilizing IMS data with 10-fold larger pixel size. Additionally, our approach employs an optimized noise sampling technique during the diffusion model's inference process to reduce variance in the generated images, yielding reliable and repeatable virtual staining. We believe this virtual staining method will significantly expand the applicability of IMS in life sciences and open new avenues for mass spectrometry-based biomedical research.

# Introduction

Imaging mass spectrometry (IMS)[1] is a powerful tool for spatial biology, enabling the discovery of intricate relationships between molecular distributions and key tissue features and cell types[2–4]. IMS offers *in situ* pixel-wise mapping of hundreds-to-thousands of molecular species with high chemical specificity and sensitivity. By combining mass spectrometric analysis with spatial mapping, IMS has great potential as a discovery platform for biomedical research. Although multiple surface sample approaches have been utilized for imaging, matrix-assisted laser desorption/ionization (MALDI) is one of the most common IMS technologies due to its applicability to a wide range of biomolecular classes and high spatial resolution capabilities. Commercially available MALDI IMS platforms can routinely achieve 5-30 µm pixel sizes[5–7], and custom platforms have been developed to enable imaging at cellular resolution with pixel sizes approaching 1 µm[8–10]. Briefly, MALDI IMS is performed by mounting a thin tissue section to a glass slide and coating it with a UV-absorbing chemical matrix, which assists with the desorption and ionization of endogenous molecules during laser irradiation. The sample stage is scanned from one location to another, recording a mass spectrum at every pixel. An intensity heat map is then plotted across the measurement region resulting in an ion image for each molecule detected. MALDI IMS spatial resolution is determined by several factors, including the laser diameter at the sample surface, the precision of the stage movement between the pixels (*i.e.,* pitch), and the extent of molecular delocalization during the sample preparation[5,11,12].

While MALDI IMS provides broad molecular coverage and high chemical specificity, it lacks inherent histological context, making it challenging to directly link molecular profiles to precise cellular features without additional information. The primary reasons for this are (i) the relatively low spatial resolution of IMS, and (ii) the absence of color and cellular morphological contrast that experts are familiar with. Both limitations can be addressed by multimodal methods integrating IMS data with optical microscopy, often requiring advanced experimental and computational methods. For example, IMS-Microscopy integrative methods have been developed to enhance human interpretability of IMS data through applications such as spatial sharpening of IMS data[13], out-of-sample prediction of molecular distributions[14], mining of IMS-derived molecular profiles based on microscopy features[15], and others[16–18]. Each of these examples requires the collection of complimentary optical microscopy data such as histochemically stained brightfield or immunofluorescence images. However, in cases where chemical staining is necessary, it can introduce additional experimental complexity, posing a significant limitation, as it precludes the preservation of the unlabeled tissue, thereby hindering any further molecular analysis on the same sample. Although serial tissue sections can be used, this requires complicated co-registration processes to account for tissue heterogeneity and cellular differences between adjacent sections. Therefore, a technique capable of predicting high-resolution cellular histopathology cues on the basis of low-resolution IMS data is urgently needed to streamline workflows and improve molecular histology.

Over the past decade, generative AI models have made significant strides, finding wide-ranging applications for vision analysis. In the biomedical domain, one of the most notable uses of these models has been the virtual histological staining of label-free tissues, where generative models are trained to transform microscopic images of label-free tissues into their histochemically stained counterparts[19–30]. Most of these approaches rely on label-free imaging modalities that exhibit a spatial resolution comparable to the resolution of the images of the stained tissue captured by a digital histology scanner, which serves as the ground truth from the perspective of the clinical workflow. Recent efforts have also generated virtual histopathology images using segmentation masks[31], domain-specific knowledge and tissue genomics[32].

In this work, we introduce a diffusion model-based virtual histological staining technique that transforms IMS-measured ion images reporting molecular species' distributions in label-free tissue samples into super-resolved brightfield microscopy images, closely matching their histochemically stained (HS) counterparts (as illustrated in Fig. 1(a-b)). Our approach utilizes an image-conditional diffusion model, underpinned by the Brownian bridge process[33,34] (Fig. 1(c)), which integrates the low-resolution conditional input with noise estimation through an attention-based U-Net[35] (Fig. 1(f, g, h)) that incorporates the time-step information to reconstruct the high-resolution histological image of the label-free sample. Following a one-time training effort, the diffusion-based virtual staining (VS) model was able to generate brightfield microscopy equivalent Periodic Acid-Schiff (PAS)-stained images from IMS-measured ion images of label-free human kidney tissue samples (never seen before) despite the fact that the input IMS data had a 10-fold larger pixel size. Quantitative evaluations confirmed that our VS approach effectively overcomes key limitations of IMS for histological interpretation by digitally creating high-resolution histological stain images using low-resolution IMS data without the need for chemical staining and microscopic imaging of stained tissue, also eliminating image registration steps since the virtually stained images are automatically registered with the IMS data used as the label-free input of the VS model. A board-certified pathologist was able to identify key pathological structures directly from the virtually stained PAS images, demonstrating a high degree of concordance with the features observed in the corresponding histochemically stained images and emphasizing the human interpretability of this approach. Furthermore, to mitigate the inherent high variance associated with diffusion models, we employed an optimized noise sampling strategy, which eliminates additive random noise in the final stages of the reverse diffusion process. This approach not only quantitatively reduces output variance but also ensures that the pathological features in the diffusion model outputs from different test runs remain consistent and histologically equivalent. In summary, our diffusion-based virtual staining approach overcomes some of the limitations of traditional IMS, including its relatively low spatial resolution and the absence of cellular morphological contrast, also eliminating the need for time-intensive histological staining process and complex image coregistration after the IMS data acquisition. We believe that this technique offers significant benefits for mass spectrometry-based molecular histology and will help accelerate IMS-enabled histological analyses in life sciences.

## Results

**Virtual histological staining of IMS-measured ion images of label-free tissue using a diffusion model**

The dataset used in this study comprises IMS-measured ion images of label-free human kidney tissues and high-resolution brightfield images of the histochemically stained versions of the same tissue samples. As shown in Fig. 1(b), the IMS data of each tissue sample were acquired using pixel-wise raster scanning with 10 µm lateral spacing. Subsequent data processing selected the most representative channels[36], resulting in individual ion images containing 1,453 mass-over-charge (m/z) channels, each with a pixel size of 10 µm. After the IMS data acquisition, the label-free tissue slides were subjected to histopathological PAS staining and digitally imaged using a benchtop brightfield optical microscope. The brightfield images of the histochemically stained tissue samples were then registered to their corresponding ion images, which formed our label-free input (IMS) and ground truth (brightfield and labeled) image pairs. Details of the dataset collection and preprocessing are provided in the Methods section.

The training and sampling process of the Brownian Bridge Diffusion Model (BBDM)[33] reflect the two-directional propagation of a Brownian bridge process, as depicted in Fig.1(c, d, e). The forward process starts from $x_0$, representing the target image domain of the histochemically stained brightfield images (ground truth, GT), and progresses toward $x_T$, the input image domain, corresponding to lower-resolution IMS images (with 1,453 m/z channels), processed through a shallow convolutional neural network for image dimension alignment/match, as shown in Fig. 1(b). The mean of the forward process is linearly scheduled from $x_0$ to $x_T$, while the variance evolves quadratically over time steps ($t$). In contrast, the reverse process aims to denoise the input IMS images of label-free tissue step-by-step, gradually refining the data without directly using the ground truth image $x_0$. A U-Net-based denoising neural network[33,35] is trained to estimate the posterior mean of $\boldsymbol{x}_t$ based on $\boldsymbol{x}_T$. As shown in Fig. 1(f, g, h), the denoising network is employed to consistently estimate the difference between the current state $x_t$ and the target image $x_0$ at arbitrary time steps between 0 and $T$ (see the Methods section). To improve the consistency/repeatability of the virtual staining process and minimize stochasticity in the diffusion process generated images, we implemented a deterministic noise sampling strategy alongside the standard noise sampling method. Specifically, we used a mean sampling strategy (Fig. 1(e, h)) that eliminates, after a certain time point is reached, the posterior noise introduced during the vanilla sampling process (Fig. 1(d, g)). Detailed sampling algorithms for this mean sampling strategy are provided in the Methods section.

The performance advantages of this deterministic sampling are further evaluated in the subsection "*Deterministic diffusion model inference via noise sampling engineering techniques.*"

Following the training phase, the BBDM-based VS model was tested on human kidney tissue excluded from both the training and validation datasets. Figure 2 showcases the VS results generated by the BBDM-based model using IMS data. The virtually stained outputs, as presented in Fig. 2(b), exhibit a good resemblance to the GT brightfield images, despite being generated from 10-fold larger pixel size IMS images, as illustrated in Fig. 2(a). Furthermore, a board-certified pathologist annotated key renal structures—glomeruli (denoted as G), as well as proximal (P) and distal (D) convoluted tubules—on both the VS and HS images. These structures are of clinical importance since they are involved in most renal pathological conditions. As depicted in Fig. 2(b-c), there was very good concordance in the identification of these structures between the VS and HS ground truth PAS images. This alignment was consistently demonstrated across multiple fields of view (FOVs) of tissue, underscoring the robustness and generalizability of our framework for the virtual staining of low-resolution IMS data.

To quantitatively evaluate the fidelity of the virtually generated high-resolution PAS images in replicating their HS counterparts, we conducted a comparative analysis on a test dataset comprising 36 distinct tissue FOVs, each with 640×640 pixels, all from the same patient. This evaluation focused on three critical aspects: (1) image contrast[37]; (2) image color distance and distribution analysis[38]; and (3) spatial frequency spectrum (detailed in the Methods section). These quantitative metrics reported in Fig. 3 were selected to evaluate whether the VS images generated from ion images can meaningfully contribute to the interpretation of IMS-measured tissue samples without physically performing the chemical staining and microscopic imaging of stained tissue, and, thus, effectively safeguarding the tissue for other assay types and subsequent analysis. Our approach also eliminates cumbersome image registration steps normally required to register IMS data with microscopy images of stained tissue, which is not needed here since the VS images are automatically registered with their input IMS data. As depicted in Fig. 3(a), the image contrast of the VS images displays a distribution closely aligned with their corresponding

HS brightfield images, with no statistically significant difference ($p$=0.512) as determined by a two-tailed paired *t*-test. The color distance between the VS and HS image pairs, measured according to the Commission Internationale de l'Éclairage (CIE)-94 standard, is summarized in Fig. 3(b). Since the majority of these color distances are below 1.5, the differences are barely noticeable[38], reflecting an excellent color consistency between the virtually generated images and their histochemically stained counterparts. This strong color agreement is further supported by the color distributions/histograms in the YCbCr color space across all test FOVs, as presented in Fig. 3(c).

Furthermore, to showcase the super-resolution capabilities of our VS framework, we conducted a spatial frequency spectrum analysis on the raw single-channel MS images (picked from 1,453 m/z channels based on the contrast of glomeruli), network-generated VS images, and their corresponding HS ground truth images; see the Methods section for details. This analysis, illustrated in Fig. 3(d), includes cross-sections of the radially averaged power spectra[39,40], which demonstrate an excellent match between the spatial frequency spectra of the VS and HS image pairs, as desired. These results further confirm the effectiveness of the VS output images, which successfully align with the spatial frequency spectra of the high-resolution HS images. These results demonstrate a marked improvement over the spatial frequency spectra of low-resolution MS images, underscoring the diffusion-based virtual staining framework's capacity to enhance spatial resolution significantly.

**Reduction analysis on mass spectrometry image channels**

The success of virtual staining for IMS data with a 10× super-resolution factor relies on the rich molecular information captured in ion images. To further shed light on this, we trained a series of diffusion-based VS models using different numbers of mass spectrometry (m/z) channels and evaluated their performance on the same test dataset. The IMS channel indices were selected from the top-ranking channels of the sorted list of 1,453 channels, prioritized based on signal-to-noise ratio (SNR) values. The SNR definition used in this study entails that for each IMS channel, the mean is divided by the standard deviation across all pixels within that channel. We selected four distinct sets of IMS channels from the sorted list, progressively reducing the channel count from 1,453 down to 23, which corresponds to reductions of 1-fold, 4-fold, 16-fold, and 64-fold in the total number of IMS channels utilized per input image. These selected IMS channels remained consistent throughout the training and testing of each VS model.

As shown in Fig. 4(a), reducing the number of MS channels from 1,453 to 23 resulted in a gradual degradation of the virtual staining performance, with noticeable losses in critical features such as nuclear morphology. The relationship between the number of IMS channels and VS quality was further quantified using the peak signal-to-noise ratio (PSNR) and learned perceptual image patch similarity[41] (LPIPS). Figure 4(b) presents the PSNR and LPIPS metrics for the four VS models, each trained with a different number of IMS channels, evaluated across a test dataset comprising 36 distinct tissue FOVs from a single patient, each with 640×640 pixels. As the number of used MS channels increased, the diffusion-based VS models achieved a statistically significantly higher VS fidelity, underscoring the rich molecular information present in MS data and its strong utility for label-free histological staining.

**Repeatable diffusion model inference via noise sampling engineering**

Tissue heterogeneity and inherent histochemical staining variability often lead to minor differences between adjacent tissue sections. These spatial changes usually do not influence the overall slide-level diagnosis and are well tolerated by human pathologists in their clinical workflow. However, computational models applied for tissue analysis are frequently misled by these slide-to-slide variations, resulting in lower performance. For our virtual staining model, the primary source of variation in the output VS results comes from the stochastic nature of the noise sampling process during the backward diffusion. To address this, we applied noise sampling process engineering techniques to improve VS consistency and reduce output variance without the need for fine-tuning or transfer learning on the trained model. In addition to the mean sampling strategy illustrated in Fig. 1, we also introduced an alternative "skip sampling" strategy for comparison, which directly estimates $x_0$ from the denoising network's output.

The rationale behind the mean and skip sampling strategies arises from the significant increase in the additional variance $\tilde{\delta}_t$ during the final stages of the reverse diffusion process (see Supplementary Fig. 1). Both of these sampling strategies—mean and skip—avoid this increasing noise variance in the reverse path after an engineered *exit* point $t_e$ (see the Methods section); this effectively reduces stochastic variations (observed from run to run) in the output of the diffusion model for the same label-free tissue FOV, which is highly desired for VS applications. Detailed descriptions of these engineered noise sampling techniques are provided in the Methods section and Supplementary Figs. 1-3.

To quantitatively assess the effectiveness of these noise sampling strategies, we tested the trained BBDM using the vanilla, mean, and skip sampling strategies, repeating each method five times and calculating the pixel-wise coefficient of variance (CV) across these repetitions of the diffusion-based VS process. Figure 5(a) shows the CV maps for the three distinct strategies in the YCbCr color channels, demonstrating that both the mean and skip sampling strategies effectively reduce the variance in the sampled VS images compared to the vanilla method. Additionally, we computed the average CV values across all the pixels in the test image FOVs and plotted them for each YCbCr channel, as shown in Fig. 5(b). Our results further corroborate that the mean and skip sampling strategies are effective in achieving lower output variances, indicating the repeatability of the diffusion-based VS process using these engineered noise sampling techniques.

This comparative analysis in Fig. 5 further reveals that the skip sampling strategy yields a lower average CV value compared to the mean sampling method. However, the mean sampling strategy produces results that exhibit better perceptual similarity to the ground truth histochemically stained PAS images. Additional visual and quantitative comparisons between different noise sampling strategies in diffusion-based VS models are presented in Supplementary Fig. 2. These results confirm that the mean sampling strategy is superior to both the vanilla and skip sampling strategies, achieving a lower average LPIPS as desired. Additionally, we evaluated the performance of an averaging strategy, which involves averaging independent test runs of different inferences for the same tissue FOV. While Supplementary Fig. 2(c) highlights the advantage of the averaging strategy in reducing pixel-level CV, the resulting images exhibit relatively lower contrast with a significantly worse LPIPS score, falling short of the performance achieved by the mean sampling strategy alone. Therefore, this averaging strategy would only be preferable in scenarios where high consistency is prioritized over image contrast.

The noise sampling strategies demonstrated here can be further optimized. For instance, we evaluated the performance of the mean and skip sampling strategies across eight different exit points ($t_e$), ranging from 0 (equivalent to the vanilla sampling strategy) to 100, as illustrated in Supplementary Fig. 3. Our findings indicate that both the mean and skip sampling strategies consistently outperformed the vanilla strategy across a range of exit points. Specifically, both strategies achieved optimal LPIPS performance at an exit time point $t_e \sim 10$. Consequently, selecting a well-suited evaluation metric and determining the appropriate exit point are critical for optimizing the performance of diffusion-based VS models, particularly for clinical applications.

## Discussion

The success of the presented label-free virtual staining of IMS-measured ion images, despite the 10-fold larger pixel size of the input IMS data, can be attributed to the capabilities of diffusion models in effectively capturing and modeling complex data distributions[42–44]. Historically, Generative Adversarial Network (GAN)-based approaches[45–47] have been a predominant choice in image restoration for biomedical applications, particularly in super-resolution image reconstruction tasks[40,48,49]. However, recent advancements[50,51] in the field have revealed that GANs might struggle with highly challenging image reconstruction problems, especially at extreme super-resolution factors (e.g., >8×). In contrast, diffusion models have emerged as a superior alternative, consistently producing more realistic and accurate spatial features even at these high super-resolution factors. Moreover, the complexity of *multiplexed* inference tasks that require simultaneous resolution enhancement and cross-domain image translation has further underscored the limitations of GAN-based techniques. It has been demonstrated that diffusion models outperform GANs in such multiplexed tasks[30,52], owing to their robustness in generating high-fidelity images. Additionally, the issue of mode collapse, which is a well-documented limitation of GANs when confronted with sparse or low-quality data, is well mitigated in diffusion models. Diffusion models, in general, exhibit more stable training dynamics, even when faced with significant discrepancies between the input data and ground truth images, making them more resilient in handling challenging datasets[53]. Taking these factors into account, we can conclude that the success of our VS models in generating high-resolution virtual stains from low-resolution IMS data of label-free tissue samples is a testament to the diffusion models' inherent strengths.

The repeatability and consistency of the IMS-generated virtually stained images are crucial for digital pathology interpretation and were achieved in this work using the mean and skip sampling strategies employed during the reverse process of the diffusion model. Another notable achievement of this study is its ability to enable medical experts to directly identify diagnostically relevant renal pathology structures—such as glomeruli, proximal, and distal convoluted tubules—using virtually stained images generated from lower-resolution IMS images. Historically, recognizing these structures directly from IMS images was not feasible. Conventional histology-directed IMS analysis requires both IMS data and brightfield microscopy images of the histochemically stained tissue, followed by a complex and time-consuming registration process to link the IMS data with pathology-annotated regions in optical microscopy images of stained tissue. Furthermore, histochemical staining of IMS slides prevents their utilization for genomics/epigenetics analyses and hinders IMS-molecular comparison studies. Our diffusion-based VS framework, however, enables us to bypass these limitations by offering a means of histological interpretation of IMS data for regions of interest within the kidney tissue, immediately after an IMS scan, without the need for histochemical staining. Moreover, once the diffusion model is trained, this technique can be seamlessly integrated into the post-IMS data processing pipeline without requiring any modifications to the existing hardware/setup.

We believe this transformative advancement has the potential to drive further biomedical research, particularly in the study of glomerular and tubular diseases through IMS. Furthermore, this method can augment existing IMS datasets by generating virtually stained images, offering significant advantages to the broader biomedical research community. Lastly, our reduction analyses on IMS channels further confirm the rich information encoded in IMS data and can potentially reveal the relationship of IMS channels with histochemical staining, which might prospectively facilitate a better understanding of both the virtual and histochemical staining processes.

It is also important to note that the quality of IMS-based virtual staining can be further enhanced. In this study, we demonstrated the virtual histological staining of ion images with a pixel size of ~10 μm, as this is a common spatial resolution for MALDI IMS systems. However, with recent advances in IMS technology, achieving higher spatial resolution is now possible[8–10]. Applying our VS framework to these high-resolution IMS images would undoubtedly enhance the fidelity of the virtually stained images, further elevating the precision and quality of the diffusion-model results. Furthermore, although we demonstrated the efficacy of our technique using PAS staining on human kidney samples, this label-free approach can be extended to other types of histochemical stains and various organs. This adaptability is supported by the prior success of various virtual staining techniques[19], making it versatile across different staining protocols and tissue types.

In conclusion, we demonstrated virtual histological staining using label-free, low-resolution IMS images. We firmly believe that our IMS-based virtual staining framework will become an essential tool for IMS-driven molecular pathology, effectively bridging the gap between IMS research and clinical diagnostics.

## Methods

### Sample preparation

Kidney tissue samples for this research were surgically removed during a full nephrectomy, and remnant tissue was processed for research purposes by the Cooperative Human Tissue Network at Vanderbilt University Medical Center. Participants consented to remnant tissue collection in accordance with institutional IRB policies. The study involved kidney specimens from 5 individual patients. Kidney tissue was flash-frozen over an isopentane-dry ice slurry, embedded in carboxymethylcellulose (CMC) and stored at −80 °C. The tissue was cryosectioned into 10-μm-thick sections using a CM3050 S cryostat (Leica

Biosystems, Wetzlar, Germany). The sections were then thaw-mounted onto indium tin oxide (ITO) coated glass slides (Delta Technologies, Loveland, CO) for IMS analysis or regular glass slides for histological staining. Slides were stored at -80 °C and returned to ~20 °C within a vacuum desiccator prior to further processing. To remove endogenous salt for IMS, the section was washed three times with chilled (4 °C) 150 mM ammonium formate 3 times for 45 seconds each. It was then dried with nitrogen gas to remove excess moisture.

**Autofluorescence microscopy and histochemical staining**

To enable co-registration of IMS data with histochemically stained images, we introduced autofluorescence (AF) images as an intermediate modality, to which both IMS and histochemically stained images can be registered (see details in the *Multimodal image registration* section). AF microscopy images were acquired on each tissue prior to IMS analysis using DAPI, eGFP and DsRed filters on a Zeiss AxioScan.Z1 slide scanner (Carl Zeiss Microscopy GmbH, Oberkochen, Germany). Additionally, AF was also collected after IMS data acquisition prior to matrix removal, enabling visualization of ablation marks created by the MALDI laser[16]. The resulting images have a pixel size of 0.65 μm. After the acquisition of the post-IMS autofluorescence images, the same unlabeled tissue was stained using a standard PAS staining protocol[54]. The stained tissue slides were scanned and digitized using a brightfield slide scanner (Leica Biosystems Aperio AT2). The resulting PAS-stained brightfield microscopy images have a pixel size of 0.22 μm.

**Imaging mass spectrometry**

Samples for IMS analysis were coated with 20 mg/mL solution of DAN dissolved in THF using a TM Sprayer M3 (HTX Technologies, LLC, Chapel Hill, NC, USA), yielding a 1.67 mg/cm$^2$ coating (0.05 mL/hr, 4 passes, 40 °C spray nozzle). MALDI IMS was performed on a prototype timsTOF fleX mass spectrometer (Bruker Daltonik, Bremen, Germany). The ion images were collected in negative ionization mode at 10 μm pixel size with a beam scan set to 6 μm, using 150 laser shots per pixel and 18.6% laser power (30% global attenuator and 62% local laser power) at 10 Hz. Data were acquired in negative ionization qTOF mode, covering an m/z range from 150 to 2000.

**Multimodal image registration**

The microscopy modalities were co-registered using the *elastix*[55] framework integrated into *wsireg*[56] software. The post-IMS AF was selected as the target modality to enable integration with IMS, with the PAS-stained brightfield microscopy images becoming the source modalities. The rigid and affine transformations were used since the microscopy modalities were collected on the same tissue section. All registered whole-slide images were stored in the vendor-neutral pyramidal OME-TIFF format at a common pixel size (i.e., the PAS image was resampled to the resolution of the pre-IMS AF and post-IMS AF). Furthermore, the MALDI IMS datasets were manually registered to the post-IMS AF images using the laser ablation marks and IMS pixels. The manual registration was performed using IMS Microlink software[57], where 8-12 fiducial markers were selected in both modalities to estimate the affine transformation.

**Data division and preparation**

The collected IMS data were exported from the Bruker timsTOF file format (.d) to a custom binary format for ease of access and improved performance. Each pixel/frame contains between $10^4$ and $10^5$ centroid peaks covering the entire acquisition range, which can be reconstructed into a pseudoprofile mass spectrum using Bruker's SDK (v2.21). The dataset was m/z-aligned using six internally identified peaks (appearing in at least 50% of the pixels) through the *msalign* library (v0.2.0). This step corrects spectral misalignment (drift along the m/z axis), resulting in increased overlap between spectral features (peaks) across the experiment. Subsequently, the mass axis of the data set was calibrated using the theoretical masses of the six peaks, achieving a precision of approximately ±1 ppm. Following the preprocessing steps, normalization correction factors were computed, and a total ion current (TIC) approach was used for mass spectral and ion image normalization. Subsequently, an average mass spectrum based on all pixels was calculated for each dataset. Since samples from multiple donors were used in this study, an average mass spectrum of all samples was generated and peak-picked, identifying 1,453 that were used for further analysis. The resulting average spectrum had a resolving power of ~40,000 at m/z 885.55.

To better match the dimensions of the IMS data and facilitate the transformation from IMS to histochemically stained images, we downsampled the histochemically stained images to achieve a pixel size of 1 μm, which is ten times smaller than the pixel size of IMS. The whole slide images (WSIs) of IMS and histochemically stained images, acquired from four patients, were then segmented into smaller FOV pairs of approximately 1400×1400 μm² (each IMS image with 140×140 pixels and each histochemically stained image with 1400×1400 pixels), with 10% overlap between neighboring regions. To augment the data for robust model training, the paired WSIs were spatially transformed using a combination of rotation and flipping. The transformed WSIs were segmented as described above. This process generated a training dataset containing 712 paired IMS-PAS microscopic image patches obtained from 4 de-identified patients. Additionally, 36 paired IMS-PAS microscopic image patches without data augmentation (each FOV with 64×64 pixels for IMS and 640×640 pixels for the histochemically stained image) were reserved for blind testing, obtained from a de-identified patient not included in the training set. During each training epoch, the paired image FOVs were further subdivided.

**Quantitative performance evaluation metrics**

To quantitatively evaluate the performance of PAS virtual staining results for the study reported in Figs. 3 and 4, we used 36 FOVs of virtually stained PAS images together with their corresponding histochemically stained images for paired image comparisons. In the quantitative evaluation illustrated in Fig. 3, we conducted a comprehensive comparative analysis using several features: image contrast, CIE-94[63] color distance, histogram distributions in the YCbCr color space, and spatial frequency spectrum analysis. The image contrast is defined as:

$$\text{Contrast} = \frac{A_{90\%} - A_{10\%}}{A_{90\%} + A_{10\%}}$$

where $A_{90\%}$ and $A_{10\%}$ represents the 90th percentile and 10th percentile of the intensity values in image $A$, respectively. In our case, image $A$ refers to the grey-scale version of the VS or HS image. For the color analysis, the CIE-94 color distance was calculated between the FOV-averaged color vectors of paired VS and HS images, utilizing the default parameter settings based on ref. [64]. The paired VS and HS images were then converted from RGB to YCbCr color space to facilitate a detailed comparison of the distributions/histograms in the Y, Cb, and Cr channels, performed separately. As for the spatial frequency spectrum analysis, the single channel raw MS image was bilinearly upsampled by a factor of 10, from 64×64 pixels to 640×640 pixels, matching the dimensions of the grey-scale VS and HS images. The frequency spectrum of each image was obtained by performing a two-dimensional (2D) Fourier Transform on the 10× bilinearly upsampled single-channel IMS image (selected based on the contrast of glomeruli), the VS output and the corresponding HS ground truth image. For this analysis, both the VS and HS images were processed in grey-scale. The radially averaged power spectrum was calculated according to Wang *et al.* [48]

For the evaluations reported in Fig. 4, we utilized the metrics of PSNR and LPIPS. The PSNR is defined based on mean squared error (MSE):

$$\text{MSE} = \frac{1}{MN} \sum_{m}^{M} \sum_{n}^{N} [A_{mn} - B_{mn}]^2$$

where $A\ and\ B$ present the histochemically and virtually stained brightfield PAS images, respectively. $m, n$ are the pixel indices, and $MN$ denotes the total number of pixels in each image. PSNR can be denoted as:

$$\text{PSNR} = 10\log_{10}\left(\frac{\max{(A)}^2}{\text{MSE}}\right)$$

where $\max{(A)}$ is the maximum pixel value of the ground truth histochemically stained PAS image.

The calculation of the LPIPS metrics utilized a pre-trained VGG network[65] to evaluate the learned perceptual similarity between the generated VS images $n$ and their corresponding HS images $n_0$. These compared image pairs were fed into the pre-trained VGG network and their feature stack from $L$ layers can be extracted as $\widehat{m}^l, \widehat{m}_0^l \in \mathbb{R}^{H_l \times W_l \times C_l}$ for layer $l$. The LPIPS score can be denoted as:

$$d(n, n_0) = \sum_l \frac{1}{H_l W_l} \sum_{h,w} \left\| (\hat{m}_{hw}^l - \hat{m}_{0hw}^l) \right\|_2^2$$

The Frechet inception distance (FID[66]) values, reported in Supplementary Fig. 3, are calculated as follows:

$$\text{FID} = \|\mu - \mu_\omega\|^2 + tr\left(\Sigma + \Sigma_\omega - 2(\Sigma\Sigma_\omega)^{\frac{1}{2}}\right)$$

where $N(\mu, \Sigma)$ represents the multivariate normal distribution estimated from the Inception v3 features[66,67] of the ground truth PAS-stained images and $N(\mu_\omega, \Sigma_\omega)$ represents the distribution estimated from the Inception v3 features of the generated virtually stained images. We used the 36 VS-HS image pairs (each with 640 × 640 pixels) as input to the Inception v3 network to extract 2048-dimensional feature vectors. From these extracted features, the mean vectors ($\mu, \mu_\omega$) and covariance matrices ($\Sigma, \Sigma_\omega$) were estimated.

The Naturalness Image Quality Evaluator (NIQE[68]) values reported in Supplementary Fig. 3, are calculated as:

$$D(\nu_1, \nu_2, \Sigma_1, \Sigma_2) = \sqrt{\left((\nu_1 - \nu_2)^T \left(\frac{\Sigma_1 + \Sigma_2}{2}\right)^{-1} (\nu_1 - \nu_2)\right)}$$

where $\nu_1$ and $\Sigma_1$ represent the mean vector and covariance matrix of our customized reference Multivariate Gaussian (MVG) model, which was fitted using 800 ground truth PAS image patches (each with 640 × 640 pixels) sampled from both the training and testing datasets. Meanwhile, $\nu_2$ and $\Sigma_2$ are the mean vector and covariance matrix for the MVG model of a generated virtually stained image. The MATLAB functions *fitniqe* and *niqe* were used to fit the reference MVG model and compute the final NIQE values for the virtually stained images[69].

**Statistical analysis**

In Fig. 3, a two-tailed paired *t*-test was conducted to assess whether the image contrast between the virtually stained output and its histochemically stained PAS counterpart was statistically equivalent, with a statistical significance level of 0.05. This analysis was performed on 36 paired VS and HS images. A *p*-value greater than 0.05 indicates no statistically significant difference in the image contrast between the VS images and their histochemically stained counterparts.

In Fig. 4, one-tailed *t*-tests were employed to assess whether there was a statistically significant improvement in the output performance between two models, denoted as Model A and Model B, trained with different numbers of MS channels. Specifically, these tests were conducted for each combination of the two models, where Model A was trained with approximately four times more MS channels than Model B (e.g., 1,453 vs. 363 MS channels). The *t*-tests were conducted across 36 unique FOVs, using the PSNR and LPIPS metrics calculated between the virtually stained images and their corresponding histochemically stained ground truth images. The null hypothesis posits that both models, despite differing in the number of MS channels, should yield identical mean PSNR and LPIPS values. A statistical significance level of 0.05 was used to reject the null hypothesis in favor of the alternative hypothesis, suggesting that Model A, trained with four times more MS channels, would yield a higher PSNR and lower LPIPS score compared to Model B. As shown in Fig. 4(b), the results indicate a statistically significant improvement in the performance of Model A compared to Model B, confirming that the number of MS channels is indeed critical for enhancing the output performance of virtual staining models.

**Other implementation details**

All network training and testing tasks were conducted on a desktop computer equipped with an Intel Core i9-13900K CPU, 64 GB of memory, and an NVIDIA GeForce RTX 4090 GPU. The code for training the diffusion models was developed in Python 3.9.19 using PyTorch 2.2.1.

## Acknowledgments

The Ozcan Research Group at UCLA acknowledges the support of NIH P41, The National Center for Interventional Biophotonic Technologies (NCIBT). The authors also acknowledge the NIH Common Fund, the National Institute of Diabetes and Digestive and Kidney Diseases (NIDDK), and the Office of the Director (OD) under Award Numbers

U54DK134302 and U01DK133766 (J.M.S. and R.V.) and in part by grant numbers 2021-240339 and 2022-309518 (L.G.M. and R.V.) from the Chan Zuckerberg Initiative DAF, an advised fund of Silicon Valley Community Foundation. The content is solely the responsibility of the authors and does not necessarily represent the official views of the National Institutes of Health. The authors report the use of Grammarly, and they take full responsibility for the content of this paper.## Contributions

A.O. conceived the idea, Y.Z. developed the image processing pipeline and prepared the training/validation and test datasets, Y.Z., L.H., trained/tested the neural networks. Y.Z., L.H., N.P., and Y.L. performed the result analysis and statistical studies. L.G.M, R.V.P and J.M.S contributed optical microscopy and IMS-related data and image registration methods. Y.Z., L.H. and A.O. prepared the manuscript. A.O. supervised the research.

## Supplementary Information includes:

- **Brownian bridge diffusion process**
- **Supplementary Figure 1**: Plot of the posterior variance $\widetilde{\delta}_t$ as a function of the reverse sampling step $t$
- **Supplementary Figure 2**: Comparison of virtual staining performance of mass spectrometry images using different sampling strategies
- **Supplementary Figure 3**: Optimization of the sampling exit point $t_e$ for both the mean and skip diffusion sampling strategies

# References

1. Caprioli, R. M., Farmer, T. B. & Gile, J. Molecular Imaging of Biological Samples: Localization of Peptides and Proteins Using MALDI-TOF MS. *Anal. Chem.* **69**, 4751–4760 (1997).
2. Colley, M. E., Esselman, A. B., Scott, C. F. & Spraggins, J. M. High-Specificity Imaging Mass Spectrometry. *Annu. Rev. Anal. Chem.* **17**, 1–24 (2024).
3. Djambazova, K. V., van Ardenne, J. M. & Spraggins, J. M. Advances in imaging mass spectrometry for biomedical and clinical research. *TrAC Trends Anal. Chem.* 117344 (2023).
4. Krestensen, K. K., Heeren, R. M. & Balluff, B. State-of-the-art mass spectrometry imaging applications in biomedical research. *Analyst* **148**, 6161–6187 (2023).
5. Spraggins, J. M. *et al.* High-Performance Molecular Imaging with MALDI Trapped Ion-Mobility Time-of-Flight (timsTOF) Mass Spectrometry. *Anal. Chem.* **91**, 14552–14560 (2019).
6. Neumann, E. K. *et al.* Spatial Metabolomics of the Human Kidney using MALDI Trapped Ion Mobility Imaging Mass Spectrometry. *Anal. Chem.* **92**, 13084–13091 (2020).
7. Taylor, M. J., Lukowski, J. K. & Anderton, C. R. Spatially Resolved Mass Spectrometry at the Single Cell: Recent Innovations in Proteomics and Metabolomics. *J. Am. Soc. Mass Spectrom.* **32**, 872–894 (2021).
8. Spivey, E. C., McMillen, J. C., Ryan, D. J., Spraggins, J. M. & Caprioli, R. M. Combining MALDI-2 and transmission geometry laser optics to achieve high sensitivity for ultra-high spatial resolution surface analysis. *J. Mass Spectrom.* **54**, 366–370 (2019).
9. Niehaus, M., Soltwisch, J., Belov, M. E. & Dreisewerd, K. Transmission-mode MALDI-2 mass spectrometry imaging of cells and tissues at subcellular resolution. *Nat. Methods* **16**, 925–931 (2019).
10. Bien, T., Bessler, S., Dreisewerd, K. & Soltwisch, J. Transmission-Mode MALDI Mass Spectrometry Imaging of Single Cells: Optimizing Sample Preparation Protocols. *Anal. Chem.* **93**, 4513–4520 (2021).
11. Puolitaival, S. M., Burnum, K. E., Cornett, D. S. & Caprioli, R. M. Solvent-free matrix dry-coating for MALDI imaging of phospholipids. *J. Am. Soc. Mass Spectrom.* **19**, 882–886 (2008).
12. Anderson, D. M. G. *et al.* A method to prevent protein delocalization in imaging mass spectrometry of non-adherent tissues: application to small vertebrate lens imaging. *Anal. Bioanal. Chem.* **407**, 2311–2320 (2015).
13. Van de Plas, R., Yang, J., Spraggins, J. & Caprioli, R. M. Image fusion of mass spectrometry and microscopy: a multimodality paradigm for molecular tissue mapping. *Nat. Methods* **12**, 366–372 (2015).

14. Tideman, L. E. *et al.* Automated biomarker candidate discovery in imaging mass spectrometry data through spatially localized Shapley additive explanations. *Anal. Chim. Acta* **1177**, 338522 (2021).
15. Verbeeck, N. *et al.* Connecting imaging mass spectrometry and magnetic resonance imaging-based anatomical atlases for automated anatomical interpretation and differential analysis. *Biochim. Biophys. Acta BBA-Proteins Proteomics* **1865**, 967–977 (2017).
16. Patterson, N. H. *et al.* Next Generation Histology-Directed Imaging Mass Spectrometry Driven by Autofluorescence Microscopy. *Anal. Chem.* **90**, 12404–12413 (2018).
17. Abdelmoula, W. M. *et al.* Automatic 3D Nonlinear Registration of Mass Spectrometry Imaging and Magnetic Resonance Imaging Data. *Anal. Chem.* **91**, 6206–6216 (2019).
18. Jones, M. A. *et al.* Discovering New Lipidomic Features Using Cell Type Specific Fluorophore Expression to Provide Spatial and Biological Specificity in a Multimodal Workflow with MALDI Imaging Mass Spectrometry. *Anal. Chem.* **92**, 7079–7086 (2020).
19. Bai, B. *et al.* Deep learning-enabled virtual histological staining of biological samples. *Light Sci. Appl.* **12**, 57 (2023).
20. Rivenson, Y. *et al.* Virtual histological staining of unlabelled tissue-autofluorescence images via deep learning. *Nat. Biomed. Eng.* **3**, 466–477 (2019).
21. Borhani, N., Bower, A. J., Boppart, S. A. & Psaltis, D. Digital staining through the application of deep neural networks to multimodal multi-photon microscopy. *Biomed. Opt. Express* **10**, 1339–1350 (2019).
22. Zhang, Y. *et al.* Digital synthesis of histological stains using micro-structured and multiplexed virtual staining of label-free tissue. *Light Sci. Appl.* **9**, 1–13 (2020).
23. Li, J. *et al.* Biopsy-free in vivo virtual histology of skin using deep learning. *Light Sci. Appl.* **10**, 1–22 (2021).
24. Pradhan, P. *et al.* Computational tissue staining of non-linear multimodal imaging using supervised and unsupervised deep learning. *Biomed. Opt. Express* **12**, 2280–2298 (2021).
25. Bai, B. *et al.* Label-Free Virtual HER2 Immunohistochemical Staining of Breast Tissue using Deep Learning. *BME Front.* **2022**, 9786242 (2022).
26. Abraham, T. M. *et al.* Label-and slide-free tissue histology using 3D epi-mode quantitative phase imaging and virtual hematoxylin and eosin staining. *Optica* **10**, 1605–1618 (2023).
27. Cao, R. *et al.* Label-free intraoperative histology of bone tissue via deep-learning-assisted ultraviolet photoacoustic microscopy. *Nat. Biomed. Eng.* **7**, 124–134 (2023).
28. Li, Y. *et al.* Virtual histological staining of unlabeled autopsy tissue. *Nat. Commun.* **15**, 1684 (2024).

29. Yang, X. *et al.* Virtual birefringence imaging and histological staining of amyloid deposits in label-free tissue using autofluorescence microscopy and deep learning. *Nat. Commun.* **15**, 7978 (2024).
30. Zhang, Y. *et al.* Super-resolved virtual staining of label-free tissue using diffusion models. *arXiv.org* https://arxiv.org/abs/2410.20073v1 (2024).
31. Winter, D. *et al.* Mask-guided cross-image attention for zero-shot in-silico histopathologic image generation with a diffusion model. Preprint at http://arxiv.org/abs/2407.11664 (2024).
32. Li, Z. *et al.* His-MMDM: Multi-domain and Multi-omics Translation of Histopathology Images with Diffusion Models. *medRxiv* 2024–07 (2024).
33. Li, B., Xue, K., Liu, B. & Lai, Y.-K. Bbdm: Image-to-image translation with brownian bridge diffusion models. in *Proceedings of the IEEE/CVF conference on computer vision and pattern Recognition* 1952–1961 (2023).
34. Revuz, D. & Yor, M. *Continuous Martingales and Brownian Motion*. vol. 293 (Springer Science & Business Media, 2013).
35. Ronneberger, O., Fischer, P. & Brox, T. U-net: Convolutional networks for biomedical image segmentation. in *International Conference on Medical image computing and computer-assisted intervention* 234–241 (Springer, 2015).
36. Esselman, A. B. *et al.* Microscopy-Directed Imaging Mass Spectrometry for Rapid High Spatial Resolution Molecular Imaging of Glomeruli. *J. Am. Soc. Mass Spectrom.* **34**, 1305–1314 (2023).
37. Janowczyk, A., Zuo, R., Gilmore, H., Feldman, M. & Madabhushi, A. HistoQC: An Open-Source Quality Control Tool for Digital Pathology Slides. *JCO Clin. Cancer Inform.* 1–7 (2019) doi:10.1200/CCI.18.00157.
38. Zhang, Y., Wu, Y., Zhang, Y. & Ozcan, A. Color calibration and fusion of lens-free and mobile-phone microscopy images for high-resolution and accurate color reproduction. *Sci. Rep.* **6**, 27811 (2016).
39. Rivenson, Y. *et al.* Deep learning microscopy. *Optica* **4**, 1437–1443 (2017).
40. de Haan, K., Ballard, Z. S., Rivenson, Y., Wu, Y. & Ozcan, A. Resolution enhancement in scanning electron microscopy using deep learning. *Sci. Rep.* **9**, 1–7 (2019).
41. Zhang, R., Isola, P., Efros, A. A., Shechtman, E. & Wang, O. The unreasonable effectiveness of deep features as a perceptual metric. in *Proceedings of the IEEE conference on computer vision and pattern recognition* 586–595 (2018).
42. Ho, J., Jain, A. & Abbeel, P. Denoising diffusion probabilistic models. *Adv. Neural Inf. Process. Syst.* **33**, 6840–6851 (2020).
43. Nichol, A. Q. & Dhariwal, P. Improved denoising diffusion probabilistic models. in *International conference on machine learning* 8162–8171 (PMLR, 2021).

44. Rombach, R., Blattmann, A., Lorenz, D., Esser, P. & Ommer, B. High-resolution image synthesis with latent diffusion models. in *Proceedings of the IEEE/CVF conference on computer vision and pattern recognition* 10684–10695 (2022).

45. Goodfellow, I. *et al.* Generative Adversarial Nets. in *Advances in Neural Information Processing Systems* (eds. Ghahramani, Z., Welling, M., Cortes, C., Lawrence, N. & Weinberger, K. Q.) vol. 27 (Curran Associates, Inc., 2014).

46. Mirza, M. Conditional generative adversarial nets. *ArXiv Prepr. ArXiv14111784* (2014).

47. Isola, P., Zhu, J.-Y., Zhou, T. & Efros, A. A. Image-to-image translation with conditional adversarial networks. in *Proceedings of the IEEE conference on computer vision and pattern recognition* 1125–1134 (2017).

48. Wang, H. *et al.* Deep learning enables cross-modality super-resolution in fluorescence microscopy. *Nat. Methods* **16**, 103–110 (2019).

49. de Haan, K., Rivenson, Y., Wu, Y. & Ozcan, A. Deep-learning-based image reconstruction and enhancement in optical microscopy. *Proc. IEEE* **108**, 30–50 (2019).

50. Saharia, C. *et al.* Image super-resolution via iterative refinement. *IEEE Trans. Pattern Anal. Mach. Intell.* **45**, 4713–4726 (2022).

51. Gao, S. *et al.* Implicit diffusion models for continuous super-resolution. in *Proceedings of the IEEE/CVF conference on computer vision and pattern recognition* 10021–10030 (2023).

52. Huang, L. *et al.* Multi-scale Conditional Generative Modeling for Microscopic Image Restoration. Preprint at http://arxiv.org/abs/2407.05259 (2024).

53. Wang, Z., Zheng, H., He, P., Chen, W. & Zhou, M. Diffusion-gan: Training gans with diffusion. *ArXiv Prepr. ArXiv220602262* (2022).

54. Neumann, E. *et al.* PAS Staining of Fresh Frozen or Paraffin Embedded Human Kidney Tissue. (2023).

55. Klein, S., Staring, M., Murphy, K., Viergever, M. A. & Pluim, J. P. Elastix: a toolbox for intensity-based medical image registration. *IEEE Trans. Med. Imaging* **29**, 196–205 (2009).

56. Patterson, H. NHPatterson/wsireg. (2022).

57. Patterson, H. NHPatterson/napari-imsmicrolink. (2022).

58. Shi, W. *et al.* Real-time single image and video super-resolution using an efficient sub-pixel convolutional neural network. in *Proceedings of the IEEE conference on computer vision and pattern recognition* 1874–1883 (2016).

59. Vaswani, A. Attention is all you need. *Adv. Neural Inf. Process. Syst.* (2017).

60. Hendrycks, D. & Gimpel, K. Gaussian error linear units (gelus). *ArXiv Prepr. ArXiv160608415* (2016).

61. Elfwing, S., Uchibe, E. & Doya, K. Sigmoid-weighted linear units for neural network function approximation in reinforcement learning. *Neural Netw.* **107**, 3–11 (2018).

62. Loshchilov, I. Decoupled weight decay regularization. *ArXiv Prepr. ArXiv171105101* (2017).
63. Industrial colour-difference evaluation | CIE. https://cie.co.at/publications/industrial-colour-difference-evaluation.
64. imcolordiff - Color difference based on CIE94 or CIE2000 standard - MATLAB. https://www.mathworks.com/help/images/ref/imcolordiff.html.
65. Simonyan, K. & Zisserman, A. Very Deep Convolutional Networks for Large-Scale Image Recognition. Preprint at http://arxiv.org/abs/1409.1556 (2015).
66. Szegedy, C., Vanhoucke, V., Ioffe, S., Shlens, J. & Wojna, Z. Rethinking the inception architecture for computer vision. in *Proceedings of the IEEE conference on computer vision and pattern recognition* 2818–2826 (2016).
67. Heusel, M., Ramsauer, H., Unterthiner, T., Nessler, B. & Hochreiter, S. Gans trained by a two time-scale update rule converge to a local nash equilibrium. *Adv. Neural Inf. Process. Syst.* **30**, (2017).
68. Mittal, A., Soundararajan, R. & Bovik, A. C. Making a "completely blind" image quality analyzer. *IEEE Signal Process. Lett.* **20**, 209–212 (2012).
69. niqe - Naturalness Image Quality Evaluator (NIQE) no-reference image quality score - MATLAB. https://www.mathworks.com/help/images/ref/niqe.html.

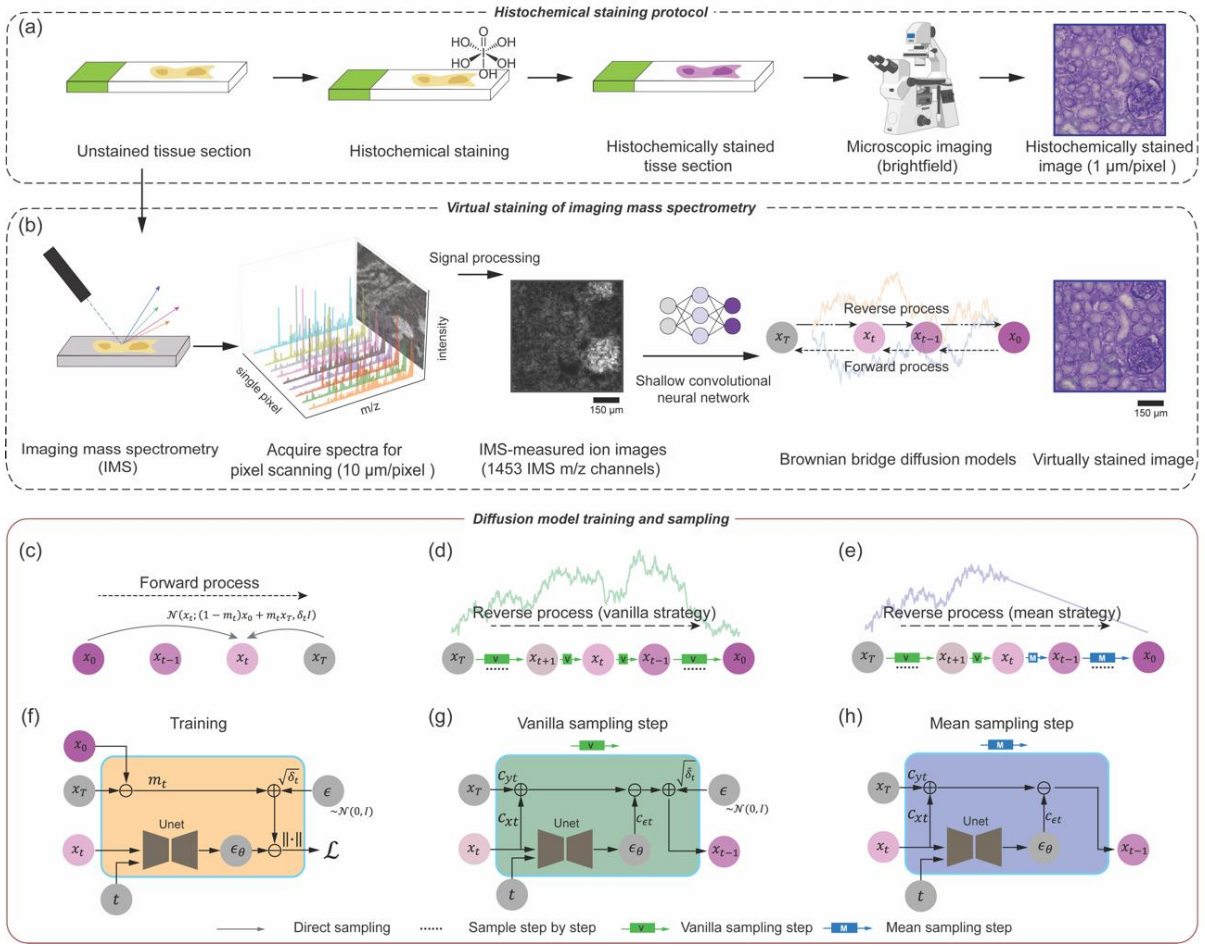

**Figure 1. Diffusion model-based virtual staining of label-free IMS-measured ion images.** (a) The workflow for histochemical staining of label-free tissue sections and the brightfield optical microscope scanning for digitization. (b) The diffusion model-based virtual staining pipeline. The Brownian bridge process is used for both the forward and reverse processes. (c) Schematic diagram of the forward process of our Brownian bridge diffusion model. (d-e) Schematic diagrams of the diffusion model reverse process that utilizes the vanilla and mean sampling strategies. (f) Detailed workflow for training the diffusion-based VS model. (g-h) Workflow for a single vanilla and mean sampling step, used individually or in combination during the reverse sampling process in (d-e).

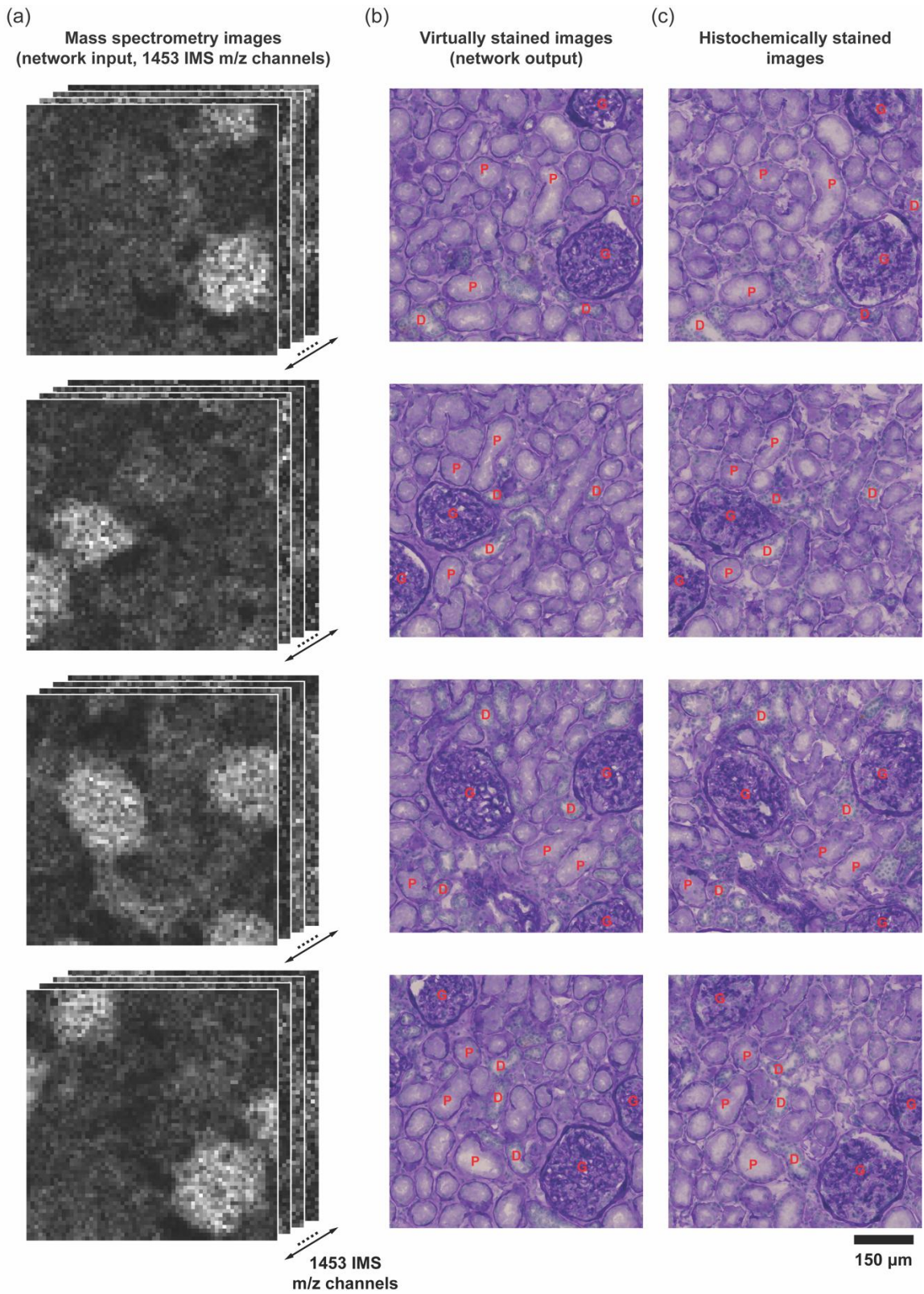

**Figure 2. Visual comparisons between the virtually stained PAS images generated from label-free IMS**

**data and their histochemically stained counterparts.** (a) Imaging mass spectrometry data of label-free tissue, consisting of 1,453 ion (m/z) channels with a pixel size of 10 μm. (b) Virtually stained images digitally generated from the IMS data using our diffusion-based VS model. (c) Histochemically stained ground truth images. Both the virtually stained and the histochemically stained images have a pixel size of 1 μm. The concordant localization of glomeruli (G), proximal convoluted tubules (P), and distal convoluted tubules (D), as annotated by a board-certified pathologist, can be visualized on both the VS and HS images.

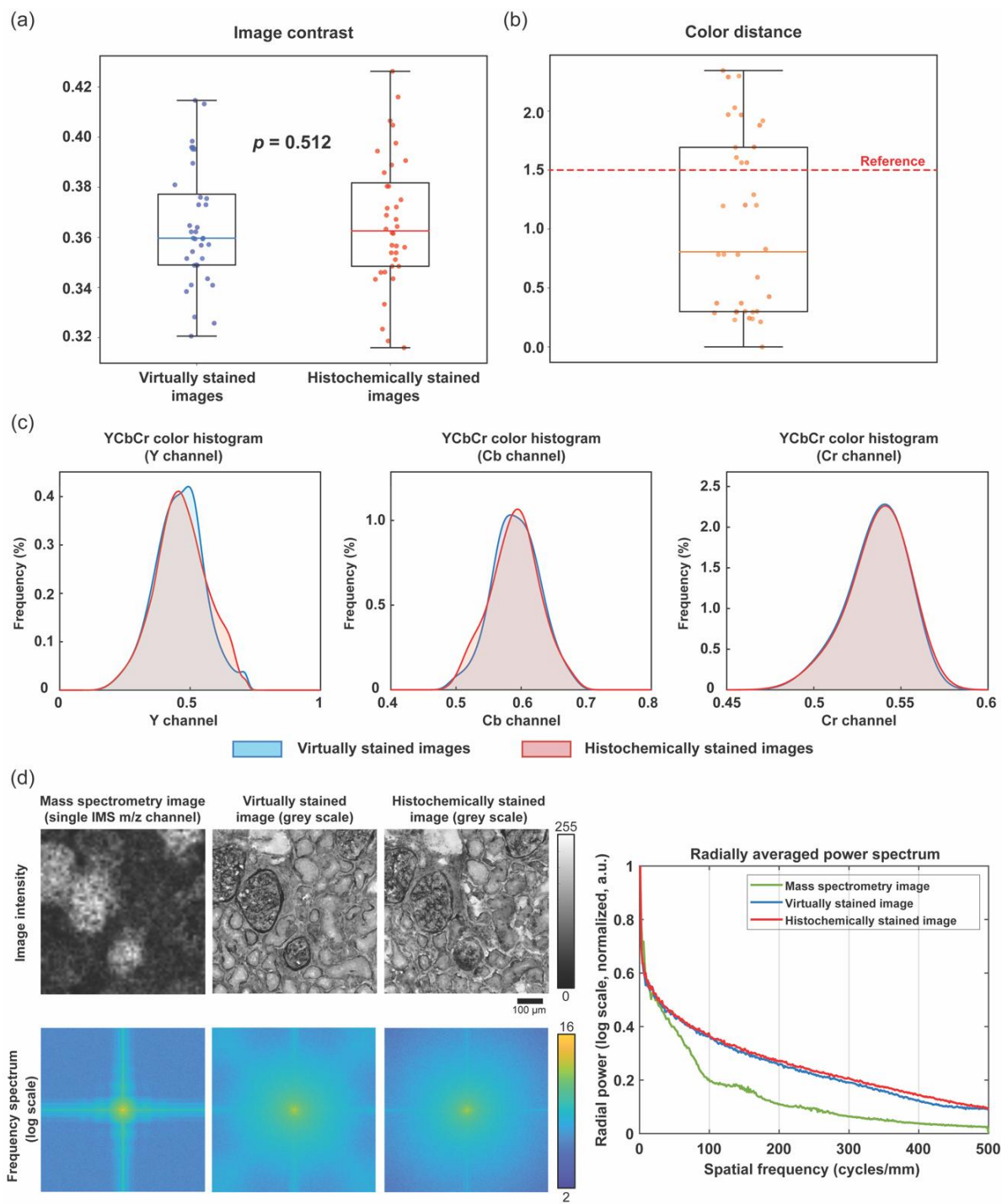

**Figure 3. Quantitative comparisons between virtually stained PAS images generated from label-free IMS data and their histochemically stained counterparts.** (a) The box plots of image contrast of virtually stained PAS images and their histochemically stained counterparts across a test dataset comprising 36 distinct FOVs. A statistical equivalence between the image contrast values of virtually and histochemically stained PAS images was determined by a two-tailed paired *t*-test (*p*=0.512). (b) The box plot of CIE-94 color distance

between the FOV-averaged color vectors of virtually stained PAS images and their corresponding histochemically stained images. (c) The color histogram comparisons of virtually stained PAS images and their histochemically stained counterparts in Y, Cb, Cr channels, calculated separately. (d) The low-resolution single MS channel image, the corresponding high-resolution grey-scale virtually and histochemically stained images (top left), and their respective spatial frequency spectra (amplitude, displayed in bottom left). The radially averaged power spectrum cross-section for each case is also presented (right).

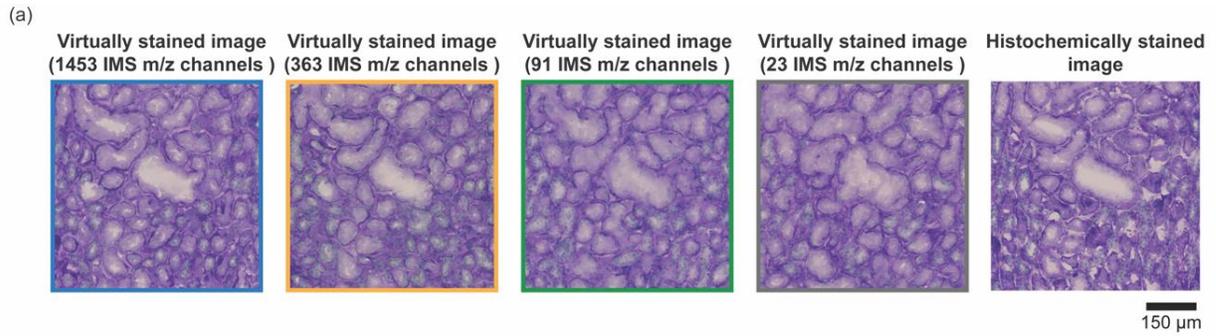

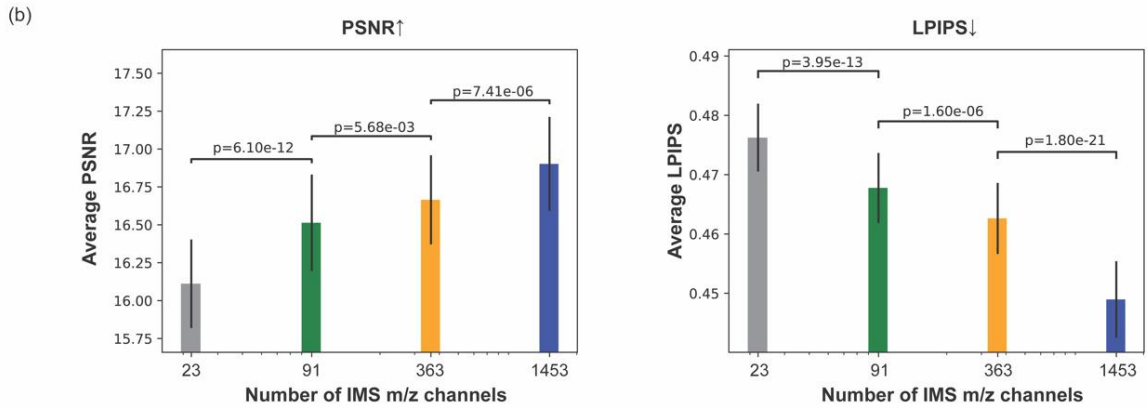

**Figure 4. Comparison of virtual staining performance for diffusion models trained with different numbers of IMS channels.** (a) Visual comparisons of virtually stained PAS images generated from diffusion-based virtual staining models trained using four different numbers of IMS channels. (b) Bar plots displaying the quantitative PSNR and LPIPS metrics averaged across testing virtually stained images for all diffusion-based virtual staining models. Higher PSNR and lower LPIPS values are desired.

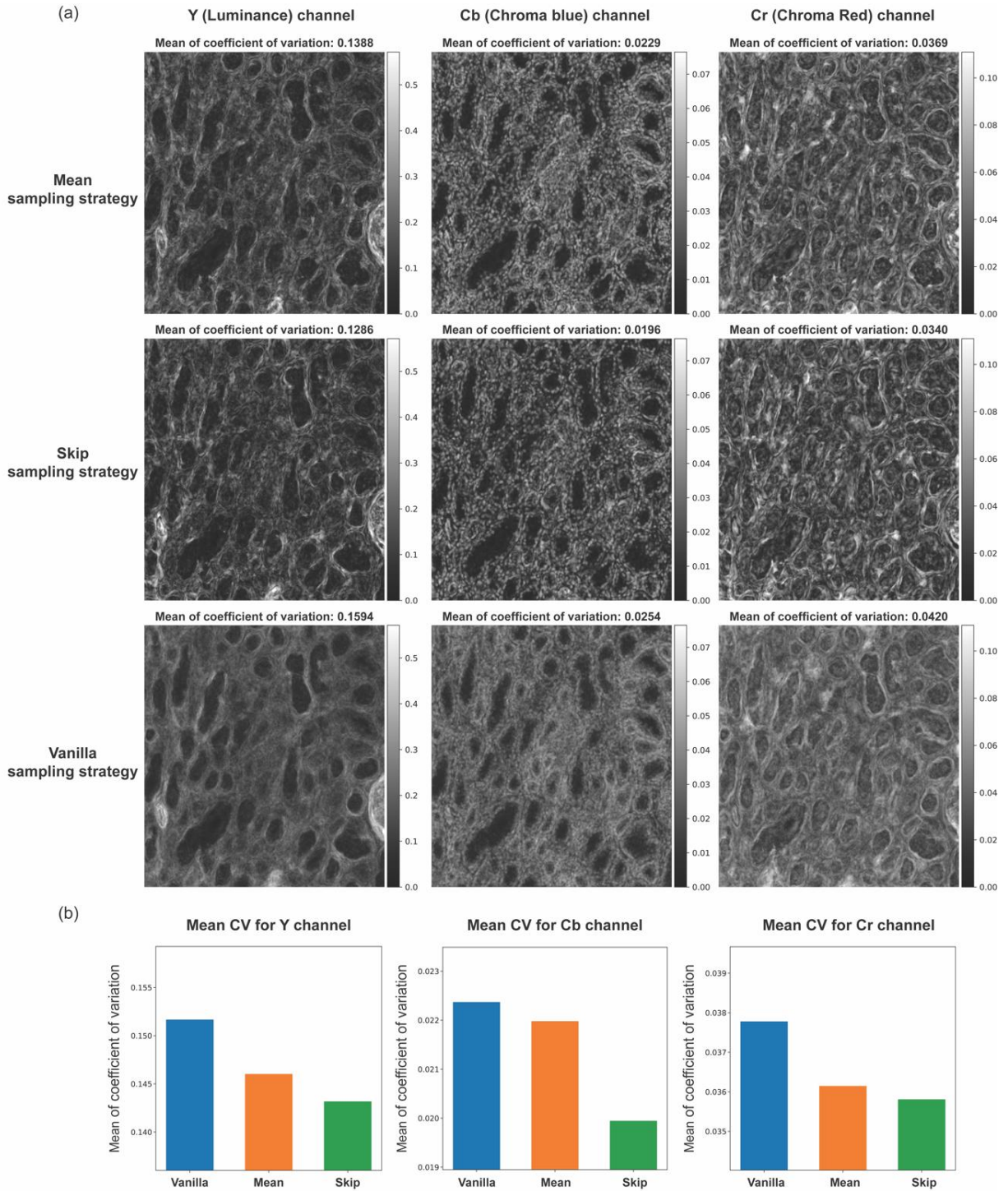

**Figure 5. Comparison of the coefficient of variation (CV) for different noise sampling engineering approaches.** (a) Visualization of the CV maps for the YCbCr channels of virtually stained images obtained using three different approaches: vanilla, mean, and skip sampling approaches. (b) Plots of the mean CV calculated across all pixels of all test image FOVs for the three sampling approaches.

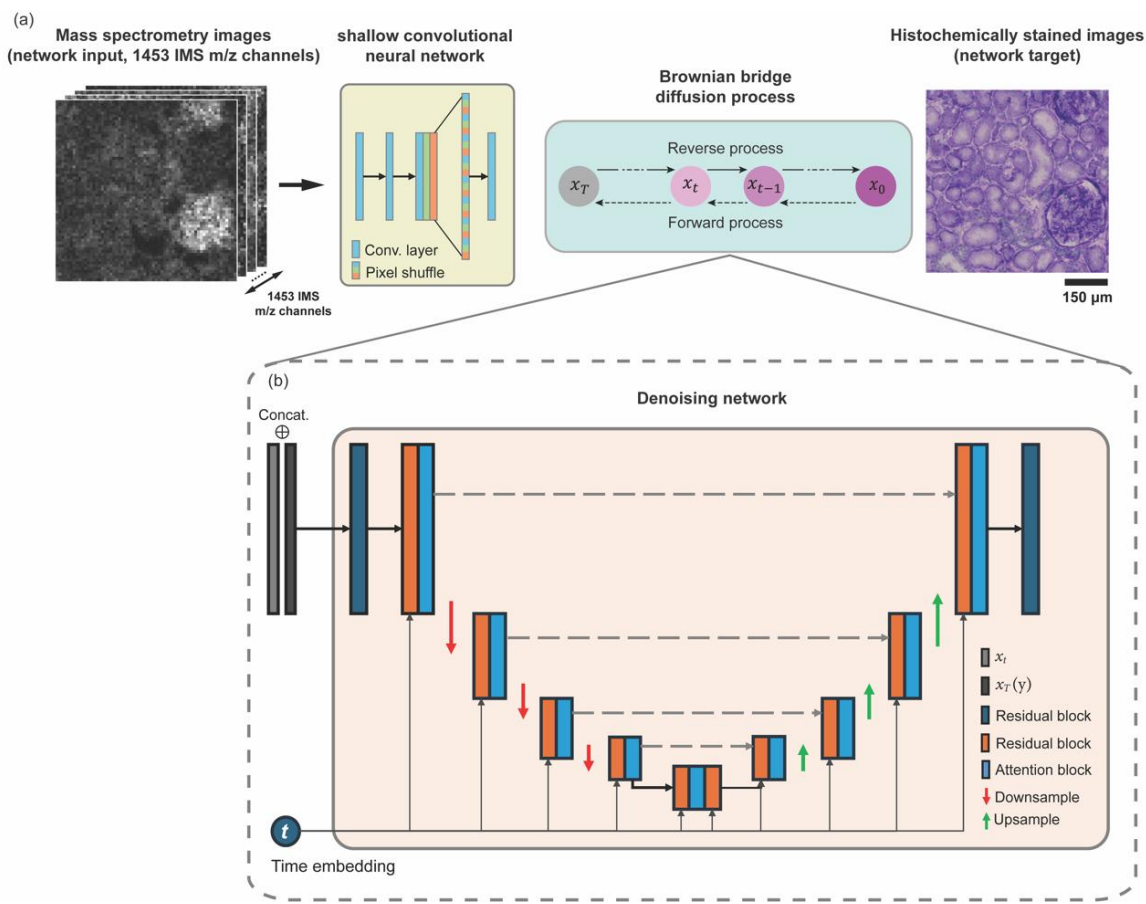

**Figure 6. The network architecture for the diffusion-based virtual staining model.** (a) The pipeline of the forward and reverse sampling processes. The detailed architecture of the shallow convolutional neural network used for dimension reduction is also presented. (b) Detailed architecture of the denoising network used at each step of both the forward and reverse sampling processes.